\documentclass[conference]{IEEEtran}
\IEEEoverridecommandlockouts
\usepackage{cite}
\usepackage{amsmath,amssymb,amsfonts}
\usepackage{algorithmic}
\usepackage{graphicx}
\usepackage{multicol}
\usepackage{tabularx}
\usepackage{steinmetz}
\usepackage{mathtools}
\usepackage{booktabs}
\usepackage{textcomp}
\usepackage{xcolor}
\usepackage{wrapfig}
\usepackage{subfig}
\usepackage{pifont}
\usepackage{tikz}
\newcommand*\circled[1]{\tikz[baseline=(char.base)]{
            \node[shape=circle,fill,inner sep=1pt] (char) {\footnotesize \textcolor{white}{#1}};}}

\usepackage[colorlinks,
            linkcolor=blue,
            anchorcolor=blue,
            citecolor=blue]{hyperref}

\newcommand{\etal}{\textit{et al.}}

\usepackage{color} 
\usepackage{tabularx} %added for tables
\usepackage{adjustbox} %added for tables
\usepackage{multirow} %added for tables
\usepackage{hyperref}

\begin{document}

\title{Towards Sparsification of Graph Neural Networks}%A Framework for Sparsifying on Temporal Graph Neural Networks

\author{
  \IEEEauthorblockN{
    Hongwu Peng\IEEEauthorrefmark{1}\textsuperscript{\textsection},
    Deniz Gurevin\IEEEauthorrefmark{1}\textsuperscript{\textsection},
    Shaoyi Huang\IEEEauthorrefmark{1},
    Tong Geng \IEEEauthorrefmark{2}, 
    Weiwen Jiang \IEEEauthorrefmark{3}, 
    Omer Khan\IEEEauthorrefmark{1}, \\
    and Caiwen Ding\IEEEauthorrefmark{1}
  }

  \IEEEauthorblockA{\IEEEauthorrefmark{1}University of Connecticut, CT, USA. \IEEEauthorrefmark{2}University of Rochester, NY, USA. \IEEEauthorrefmark{3}George Mason University, VA, USA. \\
    \IEEEauthorrefmark{1}\{hongwu.peng, deniz.gurevin, shaoyi.huang, khan, caiwen.ding\}@uconn.edu, \\
    \IEEEauthorrefmark{2}tgeng@ur.rochester.edu, 
    \IEEEauthorrefmark{3}wjiang8@gmu.edu
    \vspace{-15pt}
    }
}

\maketitle
\begingroup
\renewcommand\thefootnote{\textsection}
\footnotetext{H. Peng and D. Gurevin contributed equally to this work.}
\endgroup

\begin{abstract}
As real-world graphs expand in size, larger GNN models with billions of parameters are deployed. High parameter count in such models makes training and inference on graphs expensive and challenging. To reduce the computational and memory costs of GNNs, optimization methods such as pruning the redundant nodes and edges in input graphs have been commonly adopted. However, \textit{model compression}, which directly targets the sparsification of model layers, has been mostly limited to traditional Deep Neural Networks (DNNs) used for tasks such as image classification and object detection. In this paper, we utilize two state-of-the-art model compression methods (1) \textit{train and prune} and (2) \textit{sparse training} for the sparsification of weight layers in GNNs. We evaluate and compare the efficiency of both methods in terms of accuracy, training sparsity, and training FLOPs on real-world graphs. Our experimental results show that on the ia-email, wiki-talk, and stackoverflow datasets for link prediction, sparse training with much lower training FLOPs achieves a comparable accuracy with the train and prune method. On the brain dataset for node classification, sparse training uses a lower number FLOPs (less than 1/7 FLOPs of train and prune method) and preserves a much better accuracy performance under extreme model sparsity. Our model sparsification code is publicly available on GitHub\footnote{\url{https://github.com/HarveyP123/ICCD_SpTrn_SLR}}. 
\end{abstract}

\begin{IEEEkeywords}
graph, GNN, sparsification, model compression, sparse training, Surrogate Lagrangian Relaxation (SLR)
\end{IEEEkeywords}

% \vspace{-3mm}
% \textcolor{red}{Paper size is limited to 8 pages, including figures and references}

\section{Introduction }

%talk about GNNs  first

% \textcolor{red}{Deniz: Completed the first 3 sections. You can read and add comments if you have any feedback.}\\
% \textcolor{red}{Camera ready: 
% 1. static inputs (cora).
% 2. results of lottery ticket hypothesis
% 3. FLOPs model
% 4. future direction.}

\begin{figure}[t]
\centering
% \vspace{-3mm}
    \includegraphics[width =0.98 \linewidth]{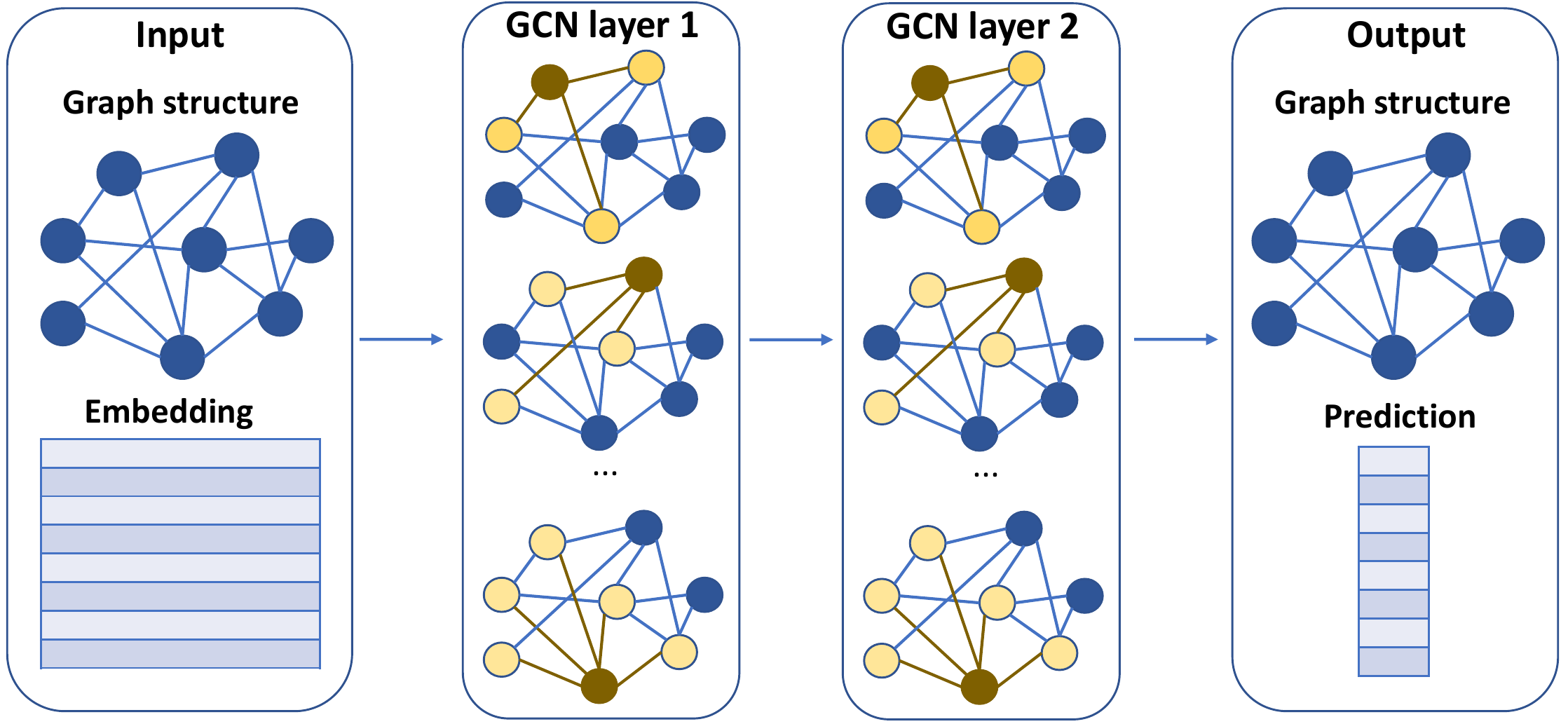}
     \caption{Graph convolution network structure.}
    %  \vspace{-5mm}
    \label{fig:GCN_structure} 
\end{figure}

Graph learning is an emerging branch in deep learning research that aims to reduce human effort in making tactical real-time decisions in applications, such as computer vision~\cite{shi2020point}, traffic forecasting~\cite{jiang2022graph}, autonomous systems~\cite{sun2019formal}, drug discovery~\cite{bongini2021molecular}, and social influence~\cite{guo2020deep}. Graph learning architectures that combine the node embeddings of a graph into neural network models have been studied and proposed, such as GNNs \cite{Scarselli2009TheGN}, Graph Convolutional Networks (GCN) \cite{Kipf2017SemiSupervisedCW}, GraphSAGE \cite{Hamilton2017InductiveRL}, and Graph Attention Networks (GAT) \cite{Velickovic2018GraphAN}. An example of GCN is given in Fig.~\ref{fig:GCN_structure}, the input of GCN is the graph structure and embedding. Each layer of GCN will aggregate the node's adjacent embedding and conduct a linear transformation with non-linear activation. The GCN final output is the prediction result for tasks.  

Increasing real-world graph sizes lead to the deployment of large GNN models with billions of parameters\cite{Sriram2022TowardsTB,manu2021co}. For example, Pinterest's PinSAGE\cite{Ying2018GraphCN} and Alibaba's AliGraph\cite{Zhu2019AliGraphAC} operate on graphs with billions of user/item embeddings (e.g., 492 million vertices, 6.8 billion edges for AliGraph). As model sizes continue to grow, GNN training has an outsize computational cost. Training such large-scale GNNs require high-end servers with expensive GPUs. that are difficult to maintain. %
%Additionally, GPUs cannot scale to process such billion-edge graphs due to their limited memory \cite{Thorpe2021DorylusAS}. 
The computational challenges of training massive GNNs with billions of parameters on large-scale graphs is an important and emerging problem in the machine learning community. Sparsifying parameters in such large GNN models can reduce the computational and memory cost in the training and inference stages.
%492.90 million vertices, 6.82 billion edges 
% Researchers estimate that it costs around \$35 million in computing power to replicate the experiments in AlphaGo Zero~\cite{mcus}. A Google study shown that GPT-3~\cite{gpt3} (with 175 billion parameters) consumed 1,287 MWh of electricity during training and produced 552 tons of carbon emissions, the equivalent of a car's 120 years of emissions~\cite{patterson2021carbon}. \textcolor{red}{Replace gpt3 and alphago with a large GNN} 

%In this paper, we apply these two categories of model sparsification methods in the context of graph learning. In our GNN sparsification framework, we sparsify the weights of a representative Feed-forward Neural Network architecture which propagates external node embeddings through its layers for link prediction and node classification tasks on graphs.

%SHOULD BE IN THE INTRO

\begin{figure}[b]
\centering
% \vspace{-3mm}
    \includegraphics[width =0.98 \linewidth]{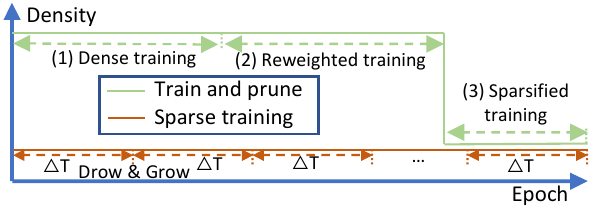}
    % \vspace{-2mm}
     \caption{Overview of two sparsification methods for neural networks: (1) \textit{train and prune} and (2) \textit{sparse training}.}
    %  \vspace{-3mm}
    \label{fig:sparsity} 
\end{figure}

There are currently two main approaches for reducing GNN training and inference complexity by sparsification: \textit{simplifying the input graph} and \textit{sparsifying the model}. The first approach, which utilizes pruning or sampling of the nodes or edges in input graphs, has been explored extensively \cite{Chen2018FastGCNFL,Yu2021GNNRLCT,Cao2019MultiChannelGN,Chen2021DyGNNAA}. 
On the other hand, while model compression (or sparsification) is well-studied for traditional Deep Neural Networks (DNNs) \cite{ding2017circnn,frankle2018lottery,kusupati2020soft, zhang2018systematic,huang2022automatic, Gurevin2021EnablingRD, peng2021optimizing, chen2021unified,tessier2022rethinking,huang2022sparse,peng2022length,qi2021accelerating,huang2021hmc,peng2021accelerating,xu2021rethinking,chen2022coarsening,li2017towards,ren2017sc}, it is an under-explored area in the context of GNNs. To the best of our knowledge, recent work by Chen \etal \cite{Chen2021AUL} was the first to propose a framework for GNNs, called Unified GNN Sparsification (UGS), that pruned the input graph as well as the model weights using the well-studied \textit{lottery ticket hypothesis} weight pruning method \cite{frankle2018lottery}. In this paper, we explore GNN sparsification using state-of-the-art model compression techniques for model weights, and evaluate their performance with both sparse \textit{and} dense node embeddings.

In general, two classes of model sparsification methods (shown in Figure \ref{fig:sparsity}) are frequently employed to achieve high scalability, performance, and energy efficiency for neural networks: (1) \textit{train and prune} (green line) and (2) \textit{sparse training} (red line). The train and prune method \cite{kusupati2020soft, zhang2018systematic, Gurevin2021EnablingRD,chen2021unified,tessier2022rethinking,huang2022sparse, peng2021binary, qi2021accommodating} first trains a dense model until it converges (step 1) and uses top-$k$ weight pruning (step 2). Since the model accuracy usually drops after this top-$k$ pruning stage, other optimization techniques, such as iterative pruning with fine-tuning and masked retraining on the sparsified model, have been employed (step 3) to recover model accuracy \cite{liu2018rethinking, singh2020woodfisher}. Model parameters are dense in the first training step
% (two third of the overall training time \cite{Gurevin2021EnablingRD}) 
and sparse only in the retraining stage (one-third of the overall training time \cite{Gurevin2021EnablingRD}). Although the final masked-retraining process is known to increase the overall run-time cost of the weight pruning pipeline \cite{Gurevin2021EnablingRD}, the initial dense training in this approach can lead to higher accuracy due to the availability of more model parameters. %has more parameters so might lead to higher accuracy

The second model sparsification method is sparse training, which starts the sparsification process of the neural network layers directly from the beginning of training, using even fewer training iterations compared to dense training. Sparse training fixes the model sparsity at the beginning of the training and uses \textit{drop and grow} policy to explore the sparse model architecture that yields the highest accuracy. It is the core of sparse training as a large number of weights are switched between zero and non-zero. Such frequent memory write and read operations heavily impact system performance. A ``proper" drop and grow policy could significantly improve the temporal and spatial locality.
Moreover, having a fixed sparsity throughout the training allows sparse training to reduce the computation and memory footprint of both training and inference stages, i.e., the weight parameters are sparse throughout the training process. The application and evaluation of the sparse training method have so far been limited only to classical DNNs for image classification tasks.

%put in the experimental section for conclusion
In this paper, we apply and compare (1) train and prune and (2) sparse training model sparsification methods in the context of graph learning. In our GNN sparsification framework, we sparsify the weights of representative Feed-forward Neural Network architecture and \textcolor{blue}{Graph Convolutional Networks (GCN)} which propagate external node embeddings through its layers for link prediction and node classification tasks on graphs. We combine and evaluate weight pruning with both dense and sparse input embeddings. In our evaluation, we compare both of the train and prune and sparse training methods in terms of accuracy, achieved sparsity, and performance. 
%\textcolor{red}{Explain experimental evaluation and findings in a paragraph...}

%we are looking at temporal graphs
In summary, our contributions are as follows
\begin{itemize}
   % \item To the best of our knowledge, this is one of the first works to apply sparsification to the weights of GNNs.
    \item We formulate two sparsification frameworks for GNNs based on (1) train and prune and (2) sparse training.
   % \item To the best of our knowledge, this is the first attempt to apply and evaluate ADMM, SLR, and sparse training-based sparsification methods on GNNs to achieve high sparsity.
    \item We evaluate and compare the trade-offs of the two evaluated sparsification methods in terms of accuracy, sparsity, and training FLOPs.
    \item To the best of our knowledge, to date, this is the first attempt that applies sparse training on graphs.
    \item For the brain, Cora, and CiteSeer dataset, we achieve a much higher accuracy using the sparse training method with much lower training FLOPs compared to the train and prune method. 
\end{itemize}

\section{Background}
% \vspace{-1mm}
\subsection{Graph Learning}

Graphs are ubiquitous data structures that describe complex systems with entities (nodes) and their interactions (edges). Large-scale and complex graph data makes machine learning tasks on graphs challenging due to having inefficient representations and requiring task-specific domain expertise. 

Graph representation learning (GRL) aims to address these challenges by encoding graph structure into a low-dimensional embedding space. GRL translates the similarity between nodes in the original graph into closeness in the embedding space. This way, graph data is represented in a lower dimensional space that reflects the underlying graph structure efficiently. 

One such GRL technique is based on performing random walks on a graph \cite{Perozzi2014DeepWalkOL, Nguyen2018ContinuousTimeDN}. Random walks capture the node properties by randomly visiting adjacent nodes. Random walks are then fed to word2vec’s skip-gram model \cite{mikolov2013efficient}, which is a natural language processing (NLP) technique, to capture node embeddings.
These node embeddings are fed into downstream graph learning tasks such as link prediction or node classification.

Graph learning is a well studied problem and there are many different techniques for it. GNNs \cite{Scarselli2009TheGN}, GCN \cite{Kipf2017SemiSupervisedCW}, GraphSAGE \cite{Hamilton2017InductiveRL}, and Graph Attention Networks (GAT) \cite{Velickovic2018GraphAN} are some of the techniques for learning inductive node embeddings that combine external node features into neural network models. 

\begin{figure}[htpb!]
\centering
% \vspace{-3mm}
    \includegraphics[width =0.98 \linewidth]{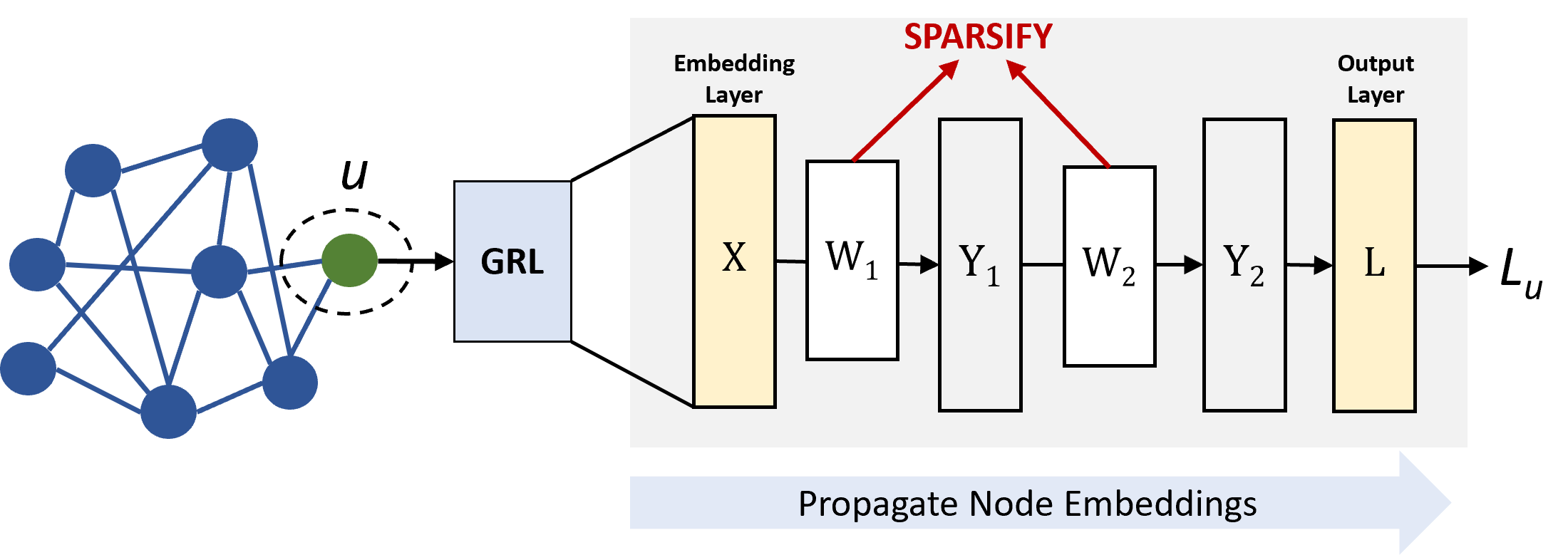}
     \caption{Sparsification of a 2-layer FNN. Given a graph, the GRL algorithm computes the embeddings of node $u$ and feeds it to the input layer of the FNN. The FNN then propagates the node embeddings through its layers to output a prediction $L_u$. We specifically focus on weight matrices sparsification.}
    \label{fig:fnn} 
    % \vspace{-3mm}
\end{figure}

Different architectures can be employed for graph prediction tasks. In this paper, we consider a commonly deployed Feed-forward Neural Network (FNN) and GCN as representative networks for graph learning. We focus on compression, or \textit{sparsification}, of the weight matrices of an FNN architecture that propagates node embeddings through its layers as shown in Figure \ref{fig:fnn}. However, our proposed sparsification framework can be applied to other GNN architectures.
%what benefits do we get by sparsifying?

% \vspace{-2mm}

\subsection{Sparsification Methods} %use it in the context of GNNs

%shorthen this paragraph
The machine learning community has recently investigated many model compression methods for Deep Neural Networks (DNNs). These methods include weight pruning, quantization, sparsity regularization, and clustering \cite{ding2017circnn,frankle2018lottery,kusupati2020soft, zhang2018systematic, Gurevin2021EnablingRD,yu2021auto,chen2021unified,tessier2022rethinking}.
The model compression techniques can reduce the learning noise and even increase the prediction accuracy \cite{krizhevsky2012advances}. The sparsified model may also increase the model robustness and has the potential to defend against adversarial attacks \cite{guo2018sparse}.

Generally, there are two major types of model sparsification methods. The first is to train the model until it converges and then prune, in which the model is pruned using top-$k$ (threshold-based) weight pruning \cite{kusupati2020soft, zhang2018systematic}. This method has also been optimized by employing iterative pruning with fine-tuning for weight dropping and accuracy retraining \cite{liu2018rethinking, singh2020woodfisher}. The second one, sparse training \cite{evci2020rigging}, gives up the hypothesis that the dense model could guide the sparsification process \cite{frankle2018lottery} and directly trains a model with fixed sparsity. % The sparse training method normally has a fixed sparsity during the entire training process and utilizes a ``drop and grow" policy to explore the proper sparse model architecture. Thus, the sparse training has better computational and memory saving than the first method. 
%Another prune during training method \cite{liu2018rethinking, singh2020woodfisher} employs prune -$>$ fine-tuning -$>$ prune iterations for weight drop and accuracy retaining, and is not the main discussion of this work. 

In this paper, we focus on these two methods of model sparsification: (1) the train and prune and (2) sparse training.

% \textcolor{red}{@Deniz}
\subsubsection{Train and Prune}

Weight pruning is one of the most common model compression methods. Several prior works have observed that a portion of weights in neural networks are redundant. Weight pruning aims to remove the redundant components in the model and achieve similar accuracy with the original model \cite{yu2021auto,chen2021unified, tessier2022rethinking,yuan2021improving}. 

Earlier work in weight pruning is mostly based on heuristic approaches \cite{He2017ChannelPF, Mao2017ExploringTR}. Later, to overcome the heuristic nature, a
systematic DNN weight pruning framework based on the Alternating Direction Methods of Multipliers (ADMM) technique \cite{Boyd2011DistributedOA} has been proposed in \cite{zhang2018systematic}. This work formulated the DNN weight pruning problem as a mathematical optimization problem and improved weight pruning by achieving $21\times$ compression on AlexNet and $71.2\times$ on LeNet-5. 

However, ADMM does not guarantee the satisfaction of all constraints because of the non-convex objective function \cite{Li2019ADMMbasedWP}. For this reason, ADMM-based weight pruning usually follows a final masked retraining process to further improve the model accuracy since the accuracy dramatically degrades after pruning. %\a typical three-stage pipeline consisting of (i) training a large model through ADMM to identify the redundant weights, (ii) hard-pruning to set the redundant weights to zero, and (iii) retraining while masking the zero-weights. 
However, the retraining phase significantly increases the overall run-time cost of the pipeline. 

To partially overcome this problem, a systematic weight pruning optimization approach based on Surrogate Lagrangian Relaxation (SLR) \cite{Bragin2015SLR} has been proposed \cite{Gurevin2021EnablingRD}. Within the SLR-based method, Lagrangian multipliers approach their optimal values faster as compared to those in the ADMM technique, and therefore, provide faster convergence during the training step and reduce final retraining cost. SLR weight pruning technique has so far been limited to classical DNNs for image classification and object detection tasks. 

\subsubsection{Sparse Training}

The train and prune method aims to reduce the computation and memory footprint at inference stage~\cite{chen2021re} (e.g., for a typical three-stage (training-pruning-retraining) pruning process, the weight parameters are dense in the first two stages (two third of the training time~\cite{Gurevin2021EnablingRD})) and are sparse in the retraining stage. Sparse training reduces the computation and memory footprint in both \textit{\textbf{training}} and \textit{\textbf{inference}} stages, i.e., the weight parameters are sparse (with fixed mask tensors) throughout the training process.

\textbf{Static mask sparse training}
% Traditional training with sparsity methods aim to reduce the computation and memory footprint at inference stage~\cite{chen2021re} (e.g., for a typical three-stage (training-pruning-retraining) pruning process, the weight parameters are dense in the first two stages (two third of the training time~\cite{gurevin2020enabling})) and sparse in the retraining stage, while static mask training reduces the computation and memory footprint to both \textit{\textbf{training}} and \textit{\textbf{inference}} stages, i.e., the weight parameters are sparse (with fixed/unchanged mask tensors) throughout the training process.
Single-Shot Network Pruning (SNIP)~\cite{lee2019snip} was the first static mask training method to train sparse sub-networks at initialization. Later, GraSP~\cite{wang2020picking} considered the weights less important if removing them would result in the least drop in the gradient norm.  Saliency criteria \cite{mozer1988skeletonization} is proposed to help decide weight importance and employed to increase the accuracy of a sparse neural network. SynFlow~\cite{tanaka2020pruning} observed that the SNIP pruning method may lead to a layer collapse phenomenon and adopts gradient-based score to avoid layer collapse. 
The mask static training methods explores unstructured sparse training, which is restricted to be deployed on hardware platform and get acceleration. Taking advantage of hardware-aware design, Pixelated Butterfly~\cite{chen2021pixelated} integrated the structured fixed sparsity butterfly format and low-rank decomposition to capture the global and local information. However, with the static masks, there is limited flexibility to preserve the important weights, so the accuracy is restricted.

% It discussed the importance of connections based on the impact on loss function at different initialization. Then for a given sparsity, redundant connections are pruned only once prior to training, and followed by the regular training.

% Pruning neural networks at initialization: Why are we missing the mark?

% % SynFlow~\cite{tanaka2020pruning},
% 3SP~\cite{van2020single}.

% Summarize all methods

% \textcolor{red}{Node:} refer to MEST~\cite{yuan2021mest} paper Section 2.2.1 Sparse Training with Fixed Sparsity Mask.

% pros: 

% cons: computation and memory-intensive

\textbf{Dynamic mask sparse training} Dynamic sparse training is the process of training with a fixed number of nonzero elements in each neural network layer. Every $\Delta T$ iteration ($\Delta T$ is the drop-and-grow frequency), a proportion of weights with least magnitude values will be dropped or set to zero, and then new weights will be randomly or greedily added to the layer in the same amount as the previously removed. Different sparse training methods usually use the same dropping method (i.e., magnitude dropping), while the growth method vary. Sparse Evolutionary Training (SET)~\cite{mocanu2018scalable} randomly grew back the previously dropped weights. RigL~\cite{evci2020rigging} grew back the weights with top-k largest gradients. SNFS~\cite{dettmers2019sparse} utilized momentum to find the important weights and layers.   ITOP~\cite{liu2021we,ma2022effective} found that the benefit of dynamic mask training come from its ability to cover all possible parameter positions. 
% \vspace{-5mm}
\section{Sparsification Frameworks for GNNs} 
% \vspace{-3mm}

In this section, we formulate the model sparsification for GNNs using (1) train and prune and (2) sparse training.
Specifically, for the train and prune, we use the SLR-based weight pruning method. For sparse training, we follow a similar drop and grow method that was proposed in RigL~\cite{evci2020rigging} to explore the sparse model architecture during the sparse training process. 
% We use a Feed Forward Neural Network (FNN) as a representative model.
% \vspace{-3mm}
%ADMM/SLR
%SPARSE TRAINING

% \vspace{-2mm}
\subsection{Weight Pruning Using SLR}
% \vspace{-2mm}

Consider a GNN with $N$ layers, where the weights at layer $n$ are denoted by ${{\bf W}}_{n}$ for $n \in \{1,2,...,N\}$. In the ADMM and SLR training, the loss function can be defined as  $ f \big( {\bf{W}}_{n}\big) +
     \sum_{n=1}^{N} g_{n}({\bf{W}}_{n})$,
for each layer $n$. The first term represents the nonlinear smooth loss function and the second term represents the non-differentiable "cardinality" penalty term~\cite{zhang2018systematic} which ensures that the number of nonzero weights are less than or equal to the predefined number $l_n$ within each layer $n$.

% \begin{eqnarray*}g_{n}({\bf{W}}_{n})=
% \begin{cases}
%  0 & \text {if } \mathrm{card}({\bf{W}}_{n})\le l_{n}, \; n = 1, \ldots, N, \\
%  +\infty & \text {otherwise. } 
% \end{cases}
% \end{eqnarray*}

%The indicator function $g_{n}(\cdot)$ ensures that the number of nonzero weights are less than or equal to the predefined number $l_n$ within each layer $n$.

The objective of SLR training is to minimize the loss function. However, because the loss function is subject to constraints on the cardinality of weights, it cannot be solved in its entirety. SLR technique decomposes the loss problem into 2 smaller subproblems by introducing duplicate variables  ${\bf{W}}_{n}={\bf{Z}}_{n}$ and rewriting the problem as $\underset{ {\bf{W}}_{n}}{\text{min}}
 f \big( {\bf{W}}_{n} \big)+\sum_{n=1}^{N} g_{n}({\bf{Z}}_{n}) \label{losscardinality}$. The \textit{Augmented} Lagrangian function ~\cite{Boyd2011DistributedOA,zhang2018systematic} of this problem can be written as:
\begin{equation}
\begin{aligned}
 L_{\rho} &   \big( {\bf{W}}_{n}, {\bf{Z}}_{n} , {\bf{\Lambda}}_{n}  \big) = 
   f \big( {\bf{W}}_{n} \big) \label{relaxedproblem}
 + \sum_{n=1}^{N} g_{n}({\bf{Z}}_{n})
\\  & +\sum_{n=1}^{N} \mathrm{tr} [{\bf{\Lambda}}_{n}^T({\bf{W}}_{n}-{\bf{Z}}_{n}) ]
 + \sum_{n=1}^{N} \frac{\rho}{2} \| {\bf{W}}_{n}-{\bf{Z}}_{n} \|_{F}^{2},
\end{aligned}
\end{equation}
where ${\bf{\Lambda}}_{n}$ are dual variables corresponding to constraints ${\bf{W}}_{n}={\bf{Z}}_{n}$. The positive scalar $\rho$ is the penalty coefficient, $\mathrm{tr}(\cdot)$ denotes the trace, and $ \| \cdot \|_{F}^{2}$ denotes the Frobenius norm. 

After the decomposition of the problem, the subproblems are solved iteratively in 2 steps: 
\begin{enumerate}
    \item Solve ``Loss Function" subproblem for ${\bf{W}}_{n}$ by using stochastic gradient descent.
    \item Solve "Cardinality" subproblem for ${\bf{Z}}_{n}$ through pruning by using projections onto discrete subspace.
\end{enumerate}

% \textbf{Alternating Direction Method of Multipliers.}
% ADMM-based weight pruning defines  ${\bf U}_{n} = (1/\rho) {\bf{\Lambda}}_{n}$. Then, the first subproblem at iteration $k$ can be equivalently written as
% \begin{equation}\label{eq:admm_loss}
% \begin{aligned}
% \underset{ {\bf{W}}_{n}}{\text{min}} \
%  f \big( {\bf{W}}_{n} \big)+\sum_{n=1}^{N} \frac{\rho}{2} \| {\bf{W}}_{n}-{\bf{Z}}_{n}^{k} + {\bf{U}}_{n}^{k} \|_{F}^{2},
%  \end{aligned}
% \end{equation}

% \noindent where the first term is the differentiable loss function of the DNN, while the other quadratic terms are differentiable and 
% convex. Therefore, the first subproblem (Eq. \ref{eq:admm_loss})
% can be solved by using stochastic gradient descent (SGD)\cite{bottou2010large}. The second subproblem is
% \begin{equation}\label{eq:admm_card}
% \begin{aligned}
% \underset{ {\bf{Z}}_{n}}{\text{min}} \
%   g_{n}({\bf{Z}}_{n}) +\sum_{n=1}^{N} \frac{\rho}{2} \| {\bf{W}}_{n}^{k+1}-{\bf{Z}}_{n} + {\bf{U}}_{n}^{k} \|_{F}^{2},
%  \end{aligned}
% \end{equation}

% The globally optimal solution
% of this can be derived using the Euclidean projection of ${\bf{W}}_{n}^{k+1}$ and ${\bf{U}}_{n}^{k}$ onto the set ${\bf{S}}_{n}= \{{\bf{W}}_{n}\mid \mathrm{card}({\bf{W}}_{n})\le l_{n}  \}, n=1,\dots,N$. After both subproblems (Eq. \ref{eq:admm_loss} and \ref{eq:admm_card}) are solved, the dual variables are updated as ${\bf{U}}_{n}^{k+1} = {\bf{U}}_{n}^{k} + {\bf{W}}_{n}^{k+1} - {\bf{Z}}_{n}^{k+1}$ 
% and this completes one iteration in ADMM regularization. 

%\textbf{Surrogate Lagrangian Relaxation.}
At iteration $k$, for given values of multipliers ${\bf{\Lambda}}_{n}^k$, the first subproblem tries to minimize the Lagrangian function, while keeping ${\bf{Z}}_{n}$ at previously obtained values ${\bf{Z}}_{n}^{k-1}$ as

\begin{equation}
\begin{aligned}
 \min_{ {\bf{W}}_{n}} L_{\rho} &   \big( {\bf{W}}_{n}, {\bf{Z}}_{n}^{k-1} , {\bf{\Lambda}}_{n} \big). \label{slr_loss}
\end{aligned}
\end{equation}
 
the subproblem can be solved by stochastic gradient descent (SGD) since the loss function of the FNN and the regularizer are differentiable . 

However, an additional "surrogate" optimality condition\cite{Bragin2015SLR} for updating the multipliers is used as follows 
  \begin{equation}
% \begin{align}
\label{SOC1}
L_{\rho} \big({\bf{W}}_{n}^{k}, {\bf{Z}}_{n}^{k-1}, {\bf{\Lambda}}_{n}^{k} \big) < 
L_{\rho} \big({\bf{W}}_{n}^{k-1}, {\bf{Z}}_{n}^{k-1}, {\bf{\Lambda}}_{n}^{k} \big)
% \nonumber
% \end{align}
\end{equation}

If this condition is satisfied, multipliers are updated as ${\bf{\Lambda}'}_{n}^{k+1}: = {\bf{\Lambda}}_{n}^{k}+ s'^k({\bf{W}}_{n}^{k}-{\bf{Z}}_{n}^{k-1})$, with an additional stepsize parameter $s_k$ \cite{Gurevin2021EnablingRD}:

\begin{equation*}
\begin{aligned}
s'^{k} =& \alpha^{k} \frac{s^{k-1}||{\bf{W}}^{k-1}-{\bf{Z}}^{k-1}||}{||{\bf{W}}^{k}-{\bf{Z}}^{k-1}||}. \label{step1a}
\end{aligned}
\end{equation*}

The stepsize parameter can be defined as $\alpha^{k} = 1 - (1 / (M \times k^{(1-\frac{1}{k^r})}))$ for
 $M > 1$, $0 < r < 1$.

The second subproblem for cardinality is solved with respect to ${\bf{Z}}_{n}$ while fixing other variables at values ${\bf{W}}_{n}^k$ as  
\begin{equation}
\begin{aligned}
 \min_{ {\bf{Z}}_{n}} L_{\rho} &   \big( {\bf{W}}_{n}^k, {\bf{Z}}_{n}, {\bf{\Lambda}'}_{n}^{k+1} \big). \label{cardinalitysubproblem}
\end{aligned}
\end{equation}

The globally optimal solution
of this can be derived using the Euclidean projection of ${\bf{W}}_{n}^{k+1}$ and ${\bf{U}}_{n}^{k}$ onto the set ${\bf{S}}_{n}= \{{\bf{W}}_{n}\mid \mathrm{card}({\bf{W}}_{n})\le l_{n}  \}, n=1,\dots,N$. This achieved through pruning using the Euclidean projection of ${\bf{W}}_{n}^{k}$ and ${\bf{\Lambda}'}_{n}^{k+1}$ onto discrete subspace. Again, in order to ensure that the multipliers' updating directions are proper, another "surrogate" optimality condition needs to be satisfied:
  \begin{equation}
% \begin{align}
\label{SOC}
 L_{\rho} \big({\bf{W}}_{n}^{k} , {\bf{Z}}_{n}^{k}, {\bf{\Lambda}'}_{n}^{k+1}  \big) < 
 L_{\rho} \big({\bf{W}}_{n}^{k} , {\bf{Z}}_{n}^{k-1} , {\bf{\Lambda}'}_{n}^{k+1} \big)
%  \nonumber
% \end{align}
\end{equation}

If this condition is not satisfied, then both subproblems are solved again by using the latest available values for  ${\bf{W}}_{n} $ and  ${\bf{Z}}_{n}$. However, if the condition is satisfied, multipliers are updated as ${\bf{\Lambda}}_{n}^{k+1}: = {\bf{\Lambda}'}_{n}^{k+1}+ s^k({\bf{W}}_{n}^{k}-{\bf{Z}}_{n}^{k})$ where the stepsizes are calculated as follows 

\begin{equation*}
\begin{aligned}
s^{k} =& \alpha^{k} \frac{s'^{k}||{\bf{W}}^{k-1}-{\bf{Z}}^{k-1}||}{||{\bf{W}}^{k}-{\bf{Z}}^{k}||}, \label{step1b}
\end{aligned}
\end{equation*}

Overall, SLR
% follows a similar approach to ADMM but additionally, it 
allows efficient subproblem solution coordination using (1) stepsizes approaching zero and (2) the satisfaction of surrogate optimality conditions ensuring updates to multipliers are assigned along correct directions. Since it supports independent and systematic adjustment of the penalty coefficient and stepsizes, model parameters obtained by SLR are much closer to their optimal values compared to ADMM, which does not support the adjustment of stepsizes without leading to slower convergence. The SLR weight pruning method has been shown to have faster convergence compared to ADMM and therefore, reduce the overall training time \cite{Gurevin2021EnablingRD}. 

After this SLR training process which adjusts the weights, the redundant weights whose values are closer to zero are pruned using Top-$k$ pruning. This can be followed by a fine-tuning step, which masks the zero weights while training, for accuracy optimizations.

\begin{figure*}[htpb!]
    %  \vspace{-.1in}
\centering
 \includegraphics[width = 0.99\linewidth]{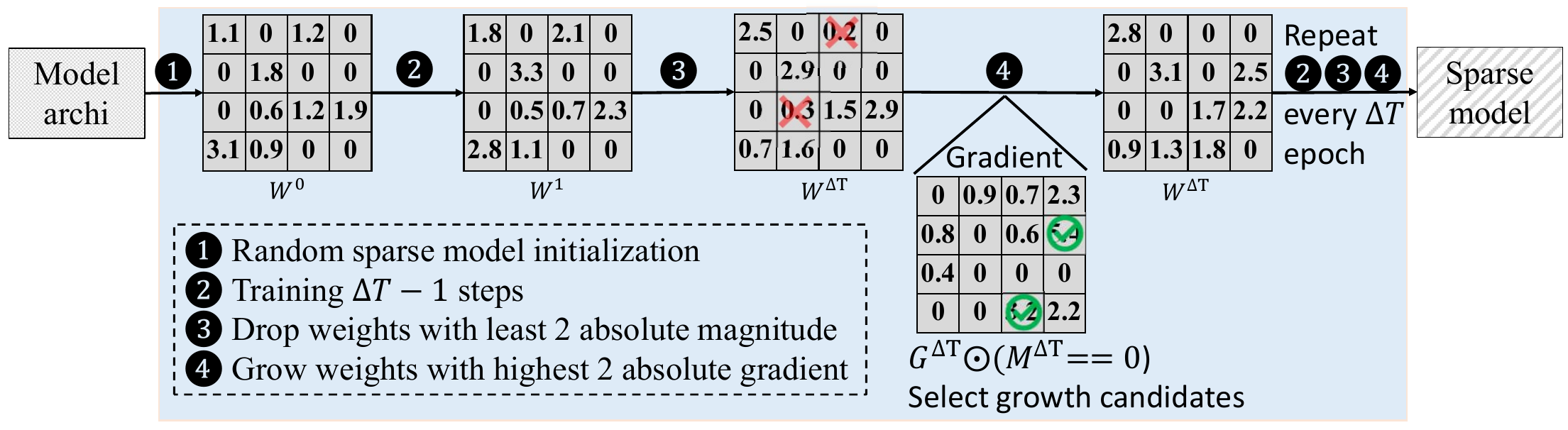}
 \centering
     \caption{Iterative drop \& grow based sparse training process.}
    %  \vspace{-7mm}
     \label{fig:sparetraining}
\end{figure*}

\subsection{Sparse Training }

\textbf{Drop-and-grow Schedule.} Dynamic sparse training is the process of training with fixed number of nonzero weights in each neural network layer.  In Fig.~\ref{fig:sparetraining}, we use a toy example to illustrate the sparse training dataflow. For simplicity, we use the matrix with a size of $4 \times 4$ to represent a weight tensor in the neural network. The sparse training is comprised of 4 steps as follows. \protect\circled{1} The weight tensor is random sparsified as $W^0$ at a given sparsity $S = 0.5$, which means 50\% of weights will be deactivated (set as zeros) and others remain activate (non-zero). \protect\circled{2} The sparsified tensor will be trained $\Delta T - 1$ iterations, where $\Delta T$ is the drop-and-grow frequency. During the $\Delta T - 1$ epochs, the non-zero elements in weight tensor are updated following the standard training process, while the zero elements will remain as zero. At the $i$-th iteration, the weight tensor is denoted as $W^i$, while the gradient is denoted as $G^i$. \protect\circled{3} At the $\Delta T$-th ietration, we first drop $k$ weights that are closed to zero or set the weights that have the least $k$ absolute magnitude as zeros ($k = 2$). Then, \protect\circled{4} we grow the weights with the highest $k$ absolute gradients back to nonzero (updating the weights with the highest $k$ absolute gradients to nonzero in the following weights updating iteration). During the process, the number of activated weights are kept the same, i.e., the newly activated (non-zero) weights are the same amount as the previously deactivated (zero) weights. \protect\circled{2}\protect\circled{3}\protect\circled{4} will be repeated till the end of the training.

\textbf{Sparse Training Forward Propagation and Back Propagation.} Consider a GNN with $L$
% fully connected 
layers. At training step $t$ and activation $a$, the collection of weights parameters of the $l$-th layer is denoted by ${\bf W}^l$, respectively.
% The output function associated with the DNN is denoted by $y=f \big({\bf W}^t, x\big)$. Weights parameters ${\bf W}^t$ are updated as ${\bf W}^{t+1} = {\bf W}^t - \eta\Delta_{{\bf W}^t} L(y,x)$, where L is the loss function during training. 
The aim of sparse training is to keep the sparsity $S$ of weight parameters as $S\in[0,1]$ during the whole training process. 
% Here $S$ is the sparsity ratio ($\alpha$).
We 
% compute the $l2$-norm of each kernel and 
drop the $\alpha$ percent of weights that are closest to zero (i.e., smallest positive weights and the largest negative weights).

% \textcolor{blue}{\textbf{Initialization}. Consider a deep neural network with $L$ layers indexed by $l \in 1,...,L$. During the training process, the weight of $i$-th layer at step $t$ are denoted by ${\bf W}_i^t$. The sparse weight tensor are randomly initialized as $\mathbf{W}^0$ with sparsity of $\theta_0$. Each sparse weight tensor within a layer has a corresponding mask tensor (zero elements masked by 0 and other elements masked by 1) with same size.} 

% \textcolor{blue}{\textbf{Training}. Zero elements in weight tensor are defined as non-active weights and others are defined as active weights. Each step, only the active weights are updated. }

% \textcolor{blue}{\textbf{Drop (deactivate):} In addition, every $\Delta T$ steps, the mask tensor is updated, i.e., the $\alpha$ percent of weights that are closest to zero (i.e., smallest positive weights and the largest negative weights) are dropped.}

% \textcolor{red}{add equation here}

% \textcolor{blue}{\textbf{Grow (activate):} Randomly growth}

% \textcolor{red}{add equation here}

% \textcolor{blue}{gradient magnitude growth}

% \textcolor{red}{add equation here}

The forward propagation could be formulated as \begin{align}
& a^l =\sigma (z^l)= \sigma (\beta^l \circledast  a^{l-1}+b^l), 
 \label{cardinalityconstraints}
\end{align}

\noindent where $z$ and $b$ represent output and biases before activation. $\circledast $ is convolution operation. $\sigma(\cdot)$ is the activation function and $a$ is the activation.
At each step, we define $\beta^l$ as a subset of weights from ${\bf W}^l$, and set the rest with zeros.
\begin{eqnarray*}\beta^t=
\begin{cases}
 {\bf W}^l & \text { if } i\in {{A}}^{l}, \\ 
 0 & \text { otherwise. }
\end{cases}
\end{eqnarray*}

\noindent where we define ${A}^{l}$ as the indices of active parameters in a sparse subset. The initial selection of ${A}^{l}$ element could be a random process~\cite{evci2020rigging} or restricted to the top-K proportion of weights by magnitude~\cite{jayakumar2020top}.

During backward propagation, we obtain the gradient of the active parameters %in a sparse subset and 
update the weights~\cite{yuan2021mest}.

% $\delta^l =\delta^{l+1}\times rotate180^\circ{\bf W}_i^{l+1}\odot \sigma^{\prime}(z^l)$,

\begin{align}
& \delta^l =\delta^{l+1}\circledast 180^\circ \text{rotation}(\beta^{l+1})\odot \sigma^{\prime}(z^l), \label{lossfunction}
\\ & G^l=a^{l-1}\circledast \delta^l \label{cardinalityconstraints}
\end{align}

\noindent where $180^\circ \text{rotation}$ represents to rotate the weight tensor $\beta^{l+1}$ 180 degrees. $\delta^l$ is the error in the $l$-th layer. $G^l$ are gradients. $\circledast$ are dot-product. $\sigma^{\prime}$ represents the derivative of activation.

% \textcolor{red}{@Hongwu Could you please add some theoretical analysis here to show that sparse training has less momery footprint requirements than dense training or train-and-prune style sparse training, based on the equations? Like weight, gradient, activation momery footprint reduction. Put the notation used in equations (6)(7)(8) into the analysis.}

\subsection{Training FLOPs Analysis}
\label{sec:train_flops}
We first evaluate the case where the input is dense and the model is dense or sparsified. We assume the forward path for a dense model has $f_D$ total number of float point operations (FLOPs) for a single epoch, and the sparsified model has $f_S$ total number of FLOPs. $f_S$ and $f_D$ can be the connected throughput sparsity $p$: $f_S = f_D * (1-p)$. Then, for each training epoch, the dense model consumes $3f_D$ FLOPs for forward and backward path \cite{evci2020rigging}, sparse model consumes $3f_S$ FLOPs for forward and backward path. For the SLR training process, assuming there are $T_1$ epochs for dense training, $T_2$ epochs for re-weight training, and $T_3$ epochs for sparsified training. The total training process of SLR training will have $T_1*f_D + T_2*f_D + T_3*f_S$ FLOPs. Assuming there are $T_s$ epochs for the sparse training process, the training process will have $T_s * f_S$  FLOPs.

% For some of the tasks, the input embedding can also be sparsified to further reduce the total number of FLOPs. However, the sparse embedding only introduces FLOPs reduction for the first layer. We assume the first layer occupies $c_1 \times$ FLOPs of a single forward epoch, and the input embedding sparsity is $p_e$. 
% % Then for the unsparsified model, a single epoch will take $(1 - p_e*c_1)*f_D$ total FLOPs during inference. For the sparsity model with sparse embedding input, a single epoch takes $(1 - p_e*c_1)*f_S$ total FLOPs. 
% In the sparse embedding case, SLR training and sparse training have $(1 - p_e*c_1)$ $\times$ total FLOPs compared to training with dense embedding. 

For some of the tasks, the input embedding can also be sparsified to reduce the total number of FLOPs further. The sparse embedding introduces a SpGEMM operation \cite{srivastava2020matraptor} into the DNN model if the weight matrix is also sparse. Assuming the sparsity of embedding is $p_{e}$ and weight sparsity if $p$. For SpGEMM operation, the probability of index matching (both locations of embedding and weight matrix have element) is $(1 - p_e) \cdot (1 - p)$. The FLOPs of the first FC layer is scaled by $(1 - p_e) \cdot (1 - p)$ times compared to the dense counterpart. Assuming the embedding dimension is $d_{emb}$, then the probability of a location of output matrix of first layer has element is $p_{o1} = (1 - (1 - p_e) \cdot (1 - p))^{d_{emb}}$, which corresponds to the sparsity of SpGEMM output matrix. With the same derivation, the second layer FLOPs is scaled by $(1 - p_{o1}) \cdot (1 - p)$ times compared to the dense case. The output sparsity of second layer is given as $p_{o2} = (1 - (1 - p_e) \cdot (1 - p))^{d_{hid2}}$, and the $d_{hid2}$ is the hidden dimension of second layer. 

\begin{figure*}[htpb!]
\centering
\includegraphics[width=0.99\linewidth]{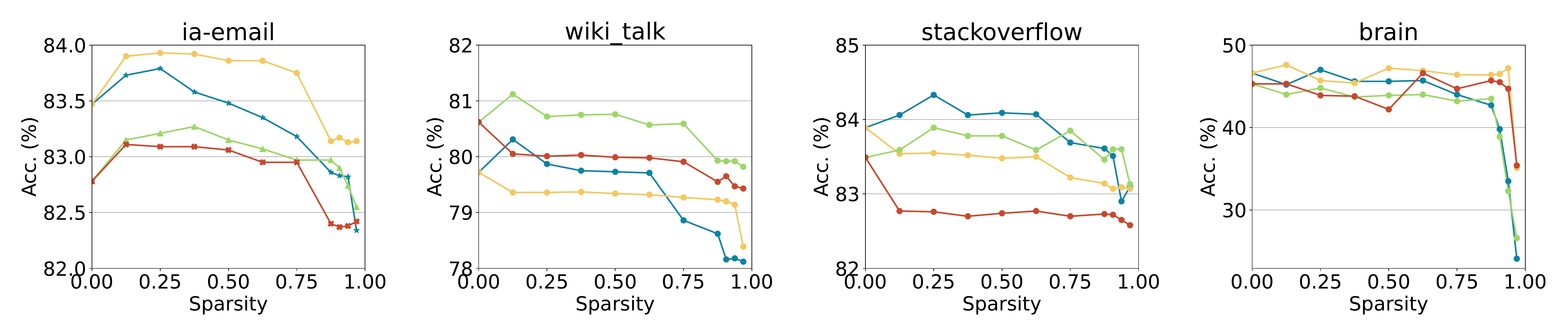}
\caption{GNN accuracy v.s. sparsity on different datasets. \textcolor[rgb]{0.04296875, 0.515625, 0.64453125}{--Blue line}: SLR \& dense embedding. \textcolor[rgb]{0.9609375, 0.78125, 0.37109375}{--Yellow line}: sparse training \& dense embedding. \textcolor[rgb]{0.61328125, 0.84375, 0.3984375}{--Green line}: SLR \& sparse embedding. \textcolor[rgb]{0.7890625, 0.27734375, 0.17578125}{--Red line}: sparse training \& sparse embedding. }
\label{fig:gnn_acc}
% \vspace{-3mm}
%\vspace{-8pt}
\end{figure*}

\section{Methodology}

\subsection{Experimental Setup}

We conduct our DNN training on an Intel Xeon Gold 5218 machine at 2.30 GHz with Ubuntu 18.04 using an Nvidia Quadro RTX 6000 GPU with 24 GB GPU memory. %\textcolor{red}{fix this if needed}

In our experiments, we firstly evaluate FNN architectures for 2 ML tasks on graphs: link prediction and node classification. We use real-world temporal graph datasets. For the link prediction task, we use wiki-talk \cite{Leskovec2014SNAPD,Paranjape2017MotifsIT,Cunningham2019CreatorGI}, ia-email \cite{Rossi2015TheND, Shetty2004TheEE} and stackoverflow \cite{Leskovec2014SNAPD,Paranjape2017MotifsIT} datasets. For the node classification task, we use brain dataset \cite{Xu2019SpatioTemporalAR, Preti2017TheDF}. The details of these datasets can be seen in Table \ref{tab:datasets}.
We also evaluate 2-layer GCN architectures with 16 hidden dimension for node classification task on graph. Cora \cite{mccallum2000automating},  Pubmed \cite{sen2008collective}, and CiteSeer \cite{giles1998citeseer} datasets are used for evaluating the GCN performance.

\begin{table}[]
\caption{Parameters of the datasets used for experiments.}
\begin{adjustbox}{width=\columnwidth}

\begin{tabular}{|c|c|c|c|}
\hline
\textbf{Task}                    & \textbf{Dataset} & \textbf{\#Nodes} & \textbf{\#Edges} \\ \hline
\multirow{3}{*}{Link Prediction} & ia-email \cite{Rossi2015TheND, Shetty2004TheEE}         & 87,274           & 1,148,072        \\ \cline{2-4} 
                                 & wiki-talk \cite{Leskovec2014SNAPD,Paranjape2017MotifsIT,Cunningham2019CreatorGI}       & 1,140,149        & 7,833,140        \\ \cline{2-4} 
                                 & stackoverflow \cite{Leskovec2014SNAPD,Paranjape2017MotifsIT}   & 6,024,271        & 63,497,050       \\ \hline
Node   Classification            & brain \cite{Xu2019SpatioTemporalAR, Preti2017TheDF}           & 5,000            & 1,955,488        \\ \hline
\end{tabular}
\label{tab:datasets}
\end{adjustbox}
% \vspace{-5mm}
\end{table}

% \textcolor{red} {@Hongwu}
%Training hardware and software used
%datasets and 
% \vspace{-1mm}
\subsection{Graph Learning}
% \vspace{-1mm}

For the graph learning tasks, we use an open-source C++ implementation\footnote{\url{https://github.com/talnish/iiswc21_rwalk}} by Talati \etal \cite{Talati2021ADD}. We use this framework for link prediction and node classification tasks on temporal graphs. 
The purpose of the node classification task is to classify a previously unseen node in a correct label/category. The link prediction task aims to predict the presence/absence of a previously unseen edge formed between 2 nodes. The link prediction is performed as a classification task: the edges present in the graph are classified as positive edges, and the edges absent in the graph are classified as negative edges.

%For each node, $15$ random walks with a length of $5$ are collected for ia-email and wiki-talk datasets and $15$ random walks of length $20$ are collected for stackoverflow dataset. These random walks are then fed to a word2vec model to produce node embeddings of size $d = 8$. 

% In order to prepare the datasets for the FNN training, the following steps are taken:
% \begin{enumerate}
%     \item $M$ temporal edges in the graph are sorted by their timestamp.
%     \item Negative sampling is used to create $M$ negative edges that do not exist in the graph.
%     \item Each edge between a pair of nodes $u$ and $v$ are represented as the concatenation of $8$ dimensional node embeddings of $u$ and $v$: $[f(u), f(v)]$ where $f(.)$ is the embedding transformation method (random walks + word2vec) producing en embedding edge of size $16$.
%     \item The positive labels are associated with label "1" and negative edges are associated with label "0".
% \end{enumerate}

 % To obtain node embeddings, $50$ random walks with a length of $15$ are collected for each node in the graph. These random walks are then fed to a word2vec model to produce node embeddings of size $d = 64$. Each node in the graph is represented by their $d$-dimensional embeddings and category label, which is a number between 0-9 in brain dataset. 

2-layer FNN with $128$ hidden dimension is used for link prediction. For node classification, a 3-layer FNN with hidden dimensions as $256$ and $128$ is used. Node embeddings are inputs for link prediction and node classification tasks. We use dense and sparse embedding for training as input. For the sparse embedding, we select the top-$k$\% values in the embedding matrix \cite{tonellotto2021query} and prune the rest. The $k$ selection considers the trade-off between embedding sparsity and information loss.  We set $k = 1.38$, $3.28$, $26.2$, $46.9$ for wiki-talk, ia-email, stackoverflow and brain datasets, respectively.   

For the 2-layer GCN on the node classification datasets, we set the hidden dimension as 16 to evaluate the performance. 
All datasets are divided as 60\%, 20\%, and 20\% for training, validation, and testing, respectively.

% This process outputs highly-dense embedding data. As well as using dense embeddings for training, we also create sparse versions of them for each dataset. We select the top-k\% values in the embedding matrix and prune the rest of the values. We use $k = $ for wiki-talk, ia-email, stackoverflow and brain datasets, respectively.
% %$k = 1.38, 3.28, 26.2, 46.9$ 

%These datasets are then used to train a 2-layer FNN for link prediction and 3-layer FNN for node classification. 

% \vspace{-1mm}
\subsection{Model Sparsification Setup}
% \vspace{-1mm}

The SLR training follows (1) the standard dense training, (2) re-weighted training through SLR and (3) sparsified training. 
The re-weighted training utilizes SLR algorithms to find the proper sparse model architecture from the dense model. After the sparse model architecture is determined, the sparsified training is applied to further retain the model accuracy for a given sparse model. 

For link prediction, we set $\rho = 0.01$, $s = 0.01$, $r = 0.1$, $M = 200$ for SLR training. For node classification, we set $\rho = 0.02$ $s = 0.02$, $r = 0.1$, $M = 200$ for SLR training. 
The initial learning rate is 0.05 for link prediction and 0.005 for node classification. We set the dense training to 20 epochs with an exponentially decayed learning rate, and the re-weight training has 10 epochs for sparse model convergence. The final sparsified training has 10 epochs. The batch size is set as 1024. 

For sparse training, we set the total number of epochs to 40 to match the total number of epochs with SLR training. We use the cosine annealing learning rate scheduler with the initial rate of $0.1$. The batch size is set as 128 for link prediction and 512 for node classification. We set the death rate as 0.5 and drop-and-grow frequency as 1000 batch iterations.
\vspace{-2mm}

% \subsection{Implementation Details}
% \textcolor{red} {@Hongwu}
% \textcolor{red} {@Shaoyi}

\section{Experimental Results}

% \textcolor{red}{Overall, sparse training can have a better performance by having less training iterations compared to dense training since it starts the sparsification directly from the beginning. Moreover, sparse training has a fixed sparsity during the entire training process, unlike the train and prune method which only trains on a sparse model in the final retraining step of a 3-stage pipeline. This allows the sparse training method to reduce the computation and memory footprint to both training and inference stages, i.e. the weight parameters are sparse (with fixed/unchanged mask tensors) throughout the training process.}

\begin{figure*}[htpb!]
\centering
\includegraphics[width=0.99\linewidth]{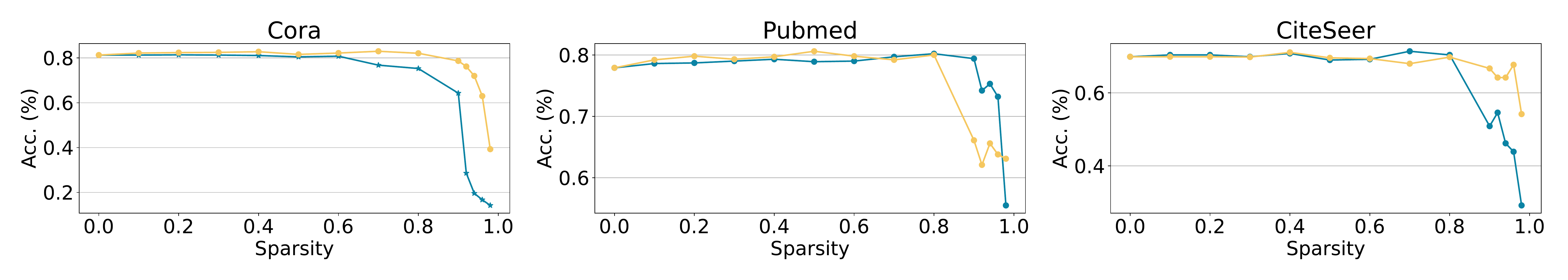}
% \vspace{-3mm}
\caption{ GCN accuracy vs. sparsity on Cora \cite{mccallum2000automating},  Pubmed \cite{sen2008collective}, and CiteSeer \cite{giles1998citeseer} datasets. \textcolor[rgb]{0.04296875, 0.515625, 0.64453125}{--Blue line}: SLR training. \textcolor[rgb]{0.9609375, 0.78125, 0.37109375}{--Yellow line}: Sparse training. Sparse embeddings are used in the experiments. } 
\label{fig:gcn_acc}
% \vspace{-6mm}
%\vspace{-8pt}
\end{figure*}

\subsection{Training FLOPs Evaluation}

We set the model parameter sparsity as 0.125, 0.25, 0.375, 0.5, 0.625, 0.75, 0.875, 0.906, 0.938, 0.969 to evaluate the performance of the SLR and sparse training method. We first evaluate the FLOPs of those two methods based on section~\ref{sec:train_flops}. By using dense training as the base, the normalized training FLOPs of SLR and sparse training can be found in Table~\ref{tab:FLOPs}. When the model sparsity is 0.906, the SLR requires more than 8 $\times$ more FLOPs than the sparse training. 

We also evaluate the embedding sparsification influence on training FLOPs in Table~\ref{tab:FLOPs}. For the link prediction tasks, the model is a 2-layer FNN, and the first layer occupies 94.1\% of total FLOPs, and thus the embedding sparsification can significantly reduce the total training FLOPs. For ia-email and wiki-talk datasets with link prediction task, the embedding sparsification brings more than 10 $\times$ training FLOPs reduction. However, for the brain dataset with node classification task, the model is 3-layer FNN, and the first layer contributes to 32.5\% of the total FLOPs. In this case, the FLOPs reduction using embedding sparsification is not significant. 

\begin{table}[htpb!]
% \vspace{-2mm}
\caption{Normalized training FLOPs (sparsity = 0.906)}
\resizebox{\linewidth}{!}
{
\begin{tabular}{|c|cc|cc|}
\hline
Training FLOPs & \multicolumn{2}{c|}{SLR}            & \multicolumn{2}{c|}{Sparse training} \\ \hline
Embedding      & \multicolumn{1}{c|}{Dense} & Sparse & \multicolumn{1}{c|}{Dense}  & Sparse \\ \hline
ia-email       & \multicolumn{1}{c|}{0.773 $\times$}  & 0.056 $\times$  & \multicolumn{1}{c|}{0.094 $\times$}   & 0.0067 $\times$  \\ \hline
wiki-talk      & \multicolumn{1}{c|}{0.773 $\times$}  & 0.069 $\times$   & \multicolumn{1}{c|}{0.094 $\times$}   & 0.0084 $\times$  \\ \hline
stackoverflow  & \multicolumn{1}{c|}{0.773 $\times$}  & 0.236 $\times$   & \multicolumn{1}{c|}{0.094 $\times$}   & 0.0286 $\times$  \\ \hline
brain          & \multicolumn{1}{c|}{0.773 $\times$}  & 0.640 $\times$   & \multicolumn{1}{c|}{0.094 $\times$}   & 0.0776 $\times$  \\ \hline
\end{tabular}
}
\label{tab:FLOPs}
% \vspace{-4mm}
\end{table}

\subsection{Accuracy Evaluation}

We further evaluate the accuracy performance of the SLR training and sparse training for different model sparsity and embedding sparsity setups. The full comparison is given in Fig.~\ref{fig:gnn_acc} and Fig.~\ref{fig:gcn_acc}. 
Most of the accuracy-sparsity curve has the Occam’s Hill \cite{rasmussen2000occam} property where the accuracy first increases with increasing sparsity and then decreases. The learned noise can be reduced with proper sparsity, which further enhances the model performance.

For both ia-email, wiki-talk, and stackoverflow datasets which are not sensitive to model parameter sparsity, sparse training has a comparable performance to the SLR method in terms of accuracy. However, for a more complex task such as node classification on the brain dataset, the sparse training will have much higher accuracy under extreme sparsity. 

For ia-email dataset, the embedding sparsification with 98.62\% decreases the model accuracy by 0.5\% with the SLR method and sparse training evaluation. As shown in Table~\ref{tab:FLOPs}, the embedding sparsification for ia-email dataset contributes to more than 10 $\times$ total FLOPs reduction for both training methods. For the wiki-talk dataset, the sparsified embedding has a positive impact on training accuracy and concurrently reduces training FLOPs more than 10 $\times$. For stackoverflow dataset, the sparse embedding reduces FLOPs more than 3 $\times$, with less than 1\% accuracy drop on average. For brain dataset with node classification task, the embedding sparsification only brings approximately 1.1 $\times$ FLOPs reduction and causes 3\% accuracy degradation on average. Thus, the embedding sparsification is not favorable for this specific task. 
For most of the evaluated datasets and tasks, the embeddings sparsification technique provides a significant training FLOPs reduction benefit and has little impact on accuracy. 

The GCN sparsification performance on Cora \cite{mccallum2000automating},  Pubmed \cite{sen2008collective}, and CiteSeer \cite{giles1998citeseer} is similar to the FNN performance. In most cases, the sparse training-based sparsification method has a better accuracy under high sparsity. The 2-layer GCN only has 16 hidden dimensions, which makes the pruning unstable under high sparsity.

\begin{figure}[htpb!]
% \vspace{-3mm}
\centering
\includegraphics[width=0.98\linewidth]{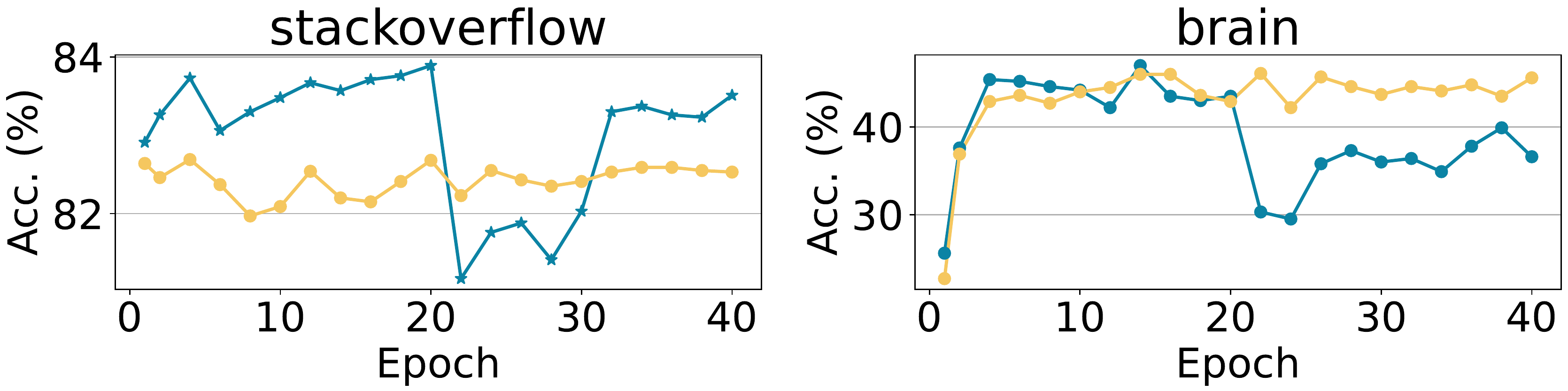}
\caption{Accuracy v.s. epoch. \textcolor[rgb]{0.04296875, 0.515625, 0.64453125}{--: SLR method}. \textcolor[rgb]{0.9609375, 0.78125, 0.37109375}{--: Sparse training}.}
\label{fig:acc_epoch}
% \vspace{-3mm}
%\vspace{-8pt}
\end{figure}

We further provide the accuracy-epoch evaluation for stackoverflow and brain datasets under 0.906 model sparsity with dense embedding. The comparison is shown in Fig.~\ref{fig:acc_epoch}. There is a significant accuracy drop at epoch 20 for the SLR algorithm as the re-weighted training starts and the parameters are pruned. For the stackoverflow dataset, the SLR-based training converges at a higher accuracy than the sparse training. However, the stackoverflow dataset has a much faster convergence rate as the accuracy remains stable for the last 20 epochs. The sparse training has a better convergence rate and accuracy than the SLR method on the brain dataset. 

%show accuracy progression through epochs
%for comparison with admm, fix one variable and show the results for other (sparsity/accuracy)

% \begin{table*}[!h]
% \small
% \centering
% \resizebox{1.0\columnwidth}{!}{
% \begin{tabular}{l|c|cccccccccccc}
% \toprule
% \multirow{2}*{\textbf{Methods}}&\multirow{2}*{\textbf{Epochs}} &  Sparsity &  Sparsity &  Sparsity&  Sparsity &  Sparsity &  Sparsity&  Sparsity &  Sparsity &  Sparsity&  Sparsity &  Sparsity &  Sparsity\\
% && 0.125 & 0.25&0.375& 0.5&0.625&0.75&0.875&0.90625&0.9375&0.96875&0.984375&0.9921875  &   &    \\
% \midrule
% \textbf{Dense} &- & & & & &79.72 && & && & & \\
% \midrule
% \multicolumn{14}{c}{\textbf{\textit{Dense wiki-talk}}} \\
% \textbf{Prune-from-dense}      &- & & & & &79.72 && & && & &  \\
% \textbf{Sparse training (ours)}      &- & & & & &79.72 && & && & &    \\
% \bottomrule
% \end{tabular}}
% \caption{Results on the graph neural network for link prediction tasks on wiki-talk~\cite{Cunningham2019CreatorGI}.}
% \label{tb:wiki-talk}
% \vspace{0.2cm}
% \end{table*}

% \subsection{Training Iterations}

%FLOPS

% \subsection{Storage Overheads} 
% %model size / parameter size
% %FLOPS
%\subsection{Computation Saving} \textcolor{red} {@Hongwu}

% \vspace{-2mm}
\section{Conclusion}

In this work, we explore the reduction of the computational and
memory costs of GNNs. We compare two types of model sparsification methods: (1) the train and prune method and (2) the sparse training method. For the train and prune, we utilize the SLR optimization method to select the proper sparse model architecture. We randomly initialize the sparse model for the sparse training approach and explore the architecture during training. The experimental results show that the sparse training method preserves much lower total FLOPs than the train and prune method, and has comparable accuracy to the train and prune method as well as a much higher accuracy under extreme sparsity. 
Sparse training method
% preserves a faster convergence rate  and
requires less training time than the SLR method. In the future, we will explore other methods for DNN sparsification, such as LTH. And we'll also explore the system speed-up of various sparsification methods for training and inference tasks.

% \vspace{3cm}

% \section{Comparison}
% Basic graph learning framework: Random Walk-Based Temporal Graph Learning

% Other graph learning frameworks: GCN, cora/ dataset

% 1. ADMM 50 for dense training (find the mask) + 50 sparse (hard mask retrain) SpMM -> SpGEMM FLOPS \\
% 2. SLR 50 for dense + 20 (hard mask retrain) sparse SpMM -$>$ SpGEMM FLOPS \\

% 2.1 SLR with un-sparsified/sparsified graph input, compare flops for pre-training and sparse (hard mask retrain) SpMM. 

% 2.2 explore different sparsity vs accuracy

% 3. Sparse train from scratch \\

% 3.1 Sparse training, explore different sparsity vs accuracy

% 3.2 similar to 2.1

% Performance: Iterations, or sparsity, or FLOPS.
% % Performance measurement: Torch.sparse supprot for SpMM is already available, how about SpGEMM (CSR format)? 

% Support for different input/dataset: cora/...

% \section*{Acknowledgment}

\bibliographystyle{IEEEtran}
\bibliography{bibligraphy}

\end{document}